\documentclass{article}

\usepackage{microtype}
\usepackage{graphicx}

\usepackage{subfigure}
\usepackage{booktabs} 

\usepackage{hyperref}

\usepackage[accepted]{icml2025}
\usepackage{booktabs} 
\usepackage{tabularx} 
\usepackage{amsmath}
\usepackage{amssymb}
\usepackage{mathtools}
\usepackage{amsthm}
\usepackage{float}
\usepackage{enumitem}
\usepackage{placeins}
\usepackage{booktabs}   
\usepackage{ragged2e}
\usepackage{multirow}   
\usepackage{colortbl}   
\usepackage{graphicx}   
\usepackage{subcaption}
\usepackage[table]{xcolor} 
\definecolor{EFE9E3}{HTML}{EFE9E3}
\definecolor{C9B59C}{HTML}{C9B59C}
\usepackage[most]{tcolorbox}
\usepackage{pifont} 
\usepackage[utf8]{inputenc} 
\makeatletter

\expandafter\let\csname algorithm*\endcsname\relax
\expandafter\let\csname endalgorithm*\endcsname\relax

\makeatother
\usepackage[ruled,vlined]{algorithm2e}
\usepackage{amsmath}
\usepackage[capitalize,noabbrev]{cleveref}

\theoremstyle{plain}

\theoremstyle{definition}

\theoremstyle{remark}

\usepackage[textsize=tiny]{todonotes}
\newcommand{\MYMETHOD}{MEME}

\icmltitlerunning{\MYMETHOD: Modeling the Evolutionary Modes of Financial Markets}

\begin{document}

\twocolumn[
\icmltitle{\textcolor{cyan}{\MYMETHOD}: \textcolor{cyan}{M}odeling the \textcolor{cyan}{E}volutionary \textcolor{cyan}{M}odes of Financial Mark\textcolor{cyan}{e}ts}



\icmlsetsymbol{leader}{$\dagger$}
\icmlsetsymbol{corr}{\ding{41}}

\begin{icmlauthorlist}
\icmlauthor{Taian Guo}{pkucs,comp}
\icmlauthor{Haiyang Shen}{pku,leader}
\icmlauthor{Junyu Luo}{pkucs}
\icmlauthor{Zhongshi Xing}{sat}
\icmlauthor{Hanchun Lian}{pkueecs}
\icmlauthor{Jinsheng Huang}{pkucs,comp}
\icmlauthor{Binqi Chen}{pkucs,comp}

\icmlauthor{Luchen Liu}{comp}
\icmlauthor{Yun Ma}{pku,corr}
\icmlauthor{Ming Zhang}{pkucs,corr}
\end{icmlauthorlist}

\icmlaffiliation{pkucs}{National Key Laboratory for Multimedia Information Processing, PKU-Anker LLM Lab, School of Computer Science, Peking University}
\icmlaffiliation{pkueecs}{School of Electronics Engineering and Computer Science, Peking University}
\icmlaffiliation{pku}{Institute for Artificial Intelligence, Peking University}
\icmlaffiliation{comp}{Zhengren Quant, Beijing, China}
\icmlaffiliation{sat}{Sun Yat-sen University}

\icmlcorrespondingauthor{Yun Ma}{mayun@pku.edu.cn}
\icmlcorrespondingauthor{Ming Zhang}{mzhang\_cs@pku.edu.cn}

\icmlkeywords{Financial Markets, Mode of Thoughts, Large Language Models}

\vskip 0.3in
]



\printAffiliationsAndNotice{\textsuperscript{$\dagger$}Project Leader. \textsuperscript{\ding{41}}Corresponding. Contact: Taian Guo \textless{}taianguo@stu.pku.edu.cn\textgreater{}, Haiyang Shen \textless{}hyshen@stu.pku.edu.cn\textgreater{}.}  

\definecolor{highlightcolor}{gray}{0.9}

\begin{abstract}
LLMs have demonstrated significant potential in quantitative finance by processing vast unstructured data to emulate human-like analytical workflows. However, current LLM-based methods primarily follow either an \textit{Asset-Centric} paradigm focused on individual stock prediction or a \textit{Market-Centric} approach for portfolio allocation, often remaining agnostic to the underlying reasoning that drives market movements. In this paper, we propose a \textit{Logic-Oriented} perspective, modeling the financial market as a dynamic, evolutionary ecosystem of competing investment narratives, termed \textbf{Modes of Thought}. To operationalize this view, we introduce \MYMETHOD~(\textcolor{cyan}{M}odeling the \textcolor{cyan}{E}volutionary \textcolor{cyan}{M}odes of Financial Mark\textcolor{cyan}{e}ts), designed to reconstruct market dynamics through the lens of evolving logics. \MYMETHOD~employs a multi-agent extraction module to transform noisy data into high-fidelity Investment Arguments and utilizes Gaussian Mixture Modeling to uncover latent consensus within a semantic space. To model semantic drift among different market conditions, we also implement a temporal evaluation and alignment mechanism to track the lifecycle and historical profitability of these modes. By prioritizing enduring market wisdom over transient anomalies, \MYMETHOD~ensures that portfolio construction is guided by robust reasoning. Extensive experiments on three heterogeneous Chinese stock pools from 2023 to 2025 demonstrate that \MYMETHOD~consistently outperforms seven SOTA baselines. Further ablation studies, sensitivity analysis, lifecycle case study and cost analysis validate \MYMETHOD's capacity to identify and adapt to the evolving consensus of financial markets. Our implementation can be found at ~\url{https://github.com/gta0804/MEME}.
\end{abstract}

\section{Introduction}

LLMs~\cite{team2025gemma, liu2025deepseek, yang2025qwen3, team2025kimi} are increasingly being applied in quantitative finance~\cite{blyth2014introduction,mcneil2015quantitative}. By processing vast amounts of unstructured data such as financial reports and social media discussions, multi-agent systems powered by LLMs can emulate the complex workflows of human analysts to distill insights and inform investment decisions. These systems demonstrate considerable promise by analyzing individual assets and adapting to changing market conditions.

Current LLM-based financial trading methods can be generally classified into two primary categories based on their analytical focus. The Asset-Centric category is primarily predictive, where the core objective is to map micro-level signals to the performance of individual stocks through forecasting~\cite{koa2024learning, zhang2024multimodal} or Alpha Factor Mining~\cite{tang2025alphaagent, shi2025navigating}. The second category, Market-Centric approaches~\cite{tong2024ploutos, guo2025mass, li2025rdagentquant}, is primarily allocative, shifting the focus to the stock pool level to optimize weights across a collection of assets. While effective, these paradigms are fundamentally object-oriented: they focus on ``what" to buy or ``how much" to allocate, while remaining agnostic to the underlying ``why". They often view the market as a static collection of data points, overlooking the latent, evolving reasoning that catalyzes market movements.

Distinct from these object-oriented paradigms, we model the financial trading problem from a Logic-Oriented perspective. We posit that the market is not merely a collection of assets, but a dynamic, evolutionary ecosystem of competing investment logics, which we term Modes of Thought. In this view, asset price movements are effectively the physical manifestations of abstract trading narratives gaining or losing consensus among market participants. However, the path to capturing these modes is obstructed by two layers of noise. First, the structural chaos and extreme noise density of multimodal data streams frequently obscure the underlying reasoning. Second, even when arguments are extracted, the reasoning itself is susceptible to logical noise—plausible-sounding narratives that are empirically invalid, contradictory, or fail to materialize in price movements. Furthermore, because individual market signals are inherently sparse and expressed through diverse terminology, the collective consensus remains hidden within idiosyncratic observations. Most crucially, the non-stationary nature of financial markets causes investment narratives to undergo constant semantic drift. Without modeling the complete lifecycle of these evolving logics and filtering out erroneous reasoning, it is hard to distinguish enduring market wisdom from fleeting, high-frequency anomalies.

To navigate these complexities, we introduce \MYMETHOD, a framework to reconstruct market dynamics through the lens of evolving Modes of Thought. Our methodology begins with a multi-agent extraction module, using specialized filtering and generation agents to transform unstructured noise into high-fidelity Investment Arguments. This collaborative architecture is essential for maintaining reasoning purity while handling vast, heterogeneous data. To uncover latent consensus, we apply Gaussian Mixture Modeling (GMM) within a semantic space. This probabilistic approach is well-suited for financial reasoning as it captures the nuanced reality where a single argument may resonate with multiple overlapping themes. To address the challenge of logical noise, \MYMETHOD~implements a temporal evaluation and alignment mechanism. By linking related modes across consecutive days and evaluating their historical profitability through exponential moving averages, the framework can distinguish between robust, value-generating market wisdom and erroneous or transient logics. This ensures that portfolio construction is guided by historically proven market narratives rather than unverified or failing reasoning.

We conduct extensive experiments on three heterogeneous stock pools in the Chinese market, including the SSE 50, CSI 300, and CSI 500, covering a volatile period from 2023 to 2025. Our evaluation compares \MYMETHOD~against eight competitive baselines, ranging from traditional gradient boosting and deep factor models to state-of-the-art LLM-based agentic frameworks. The results demonstrate that our method achieves superior profitability and stability across various market regimes. Beyond performance metrics, we provide a detailed ablation study to verify each component, a parameter sensitivity analysis, and a temporal out-of-sample test to ensure robustness against data leakage. Finally, a comprehensive case study visualizes the lifecycle of identified modes, validating the capacity of our framework to track the transition of market consensus and identify effective investment narratives as market conditions evolve.

In summary, our main contributions are as follows: 

\begin{itemize}[left=0.3cm,partopsep=-2pt,topsep=-2pt,itemsep=-2pt]
    \item We introduce \MYMETHOD, a framework that models market dynamics as an evolutionary process of competing Modes of Thought, shifting the focus from assets to the lifecycle of investment logics. 
    \item We design an end-to-end system that extracts structured arguments, identifies latent consensus via probabilistic modeling, and tracks the lifecycle of these modes through temporal alignment.
    \item We conduct extensive experiments on three stock pools with distinct styles, achieving consistent improvements over seven SOTA baselines and providing an in-depth analysis of the identified market modes.

\end{itemize}

\section{Related Work}

This section reviews literature on financial trading. We categorize existing studies into asset-centric and market-centric approaches.

\subsection{Asset-Centric Approaches}

Asset-centric research focuses on the micro-level forecasting of individual assets and the mining of predictive signals. This field includes traditional quantitative studies such as formulaic alpha factor mining and stock trend prediction~\cite{blyth2014introduction, mcneil2015quantitative}. Researchers use genetic programming~\cite{chen2021gp}, reinforcement learning~\cite{yu2023alphagen, alphaqcm}, and flow networks~\cite{chen2025alphasage} to generate predictive expressions. Other studies employ factor modeling~\cite{duan2022factorvae, xiang2024rsap, duan2025factorgcl, shi2025alphaforge} and temporal dependency analysis~\cite{yoo2021accurate, li2024master, zhang2025multi}. Advanced models also integrate entity dependencies~\cite{zhao2022stock, xu2021hist, qian2024mdgnn, song2025multi}, causality modeling~\cite{luo2023causality, li2024causalstock}, and sentiment features~\cite{araci2019finbert, du2024financial} or financial foundation models~\cite{shi2025kronos}. With the development of LLMs, agent-based methods have emerged. These systems perform multi-modal analysis~\cite{zhang2024multimodal}, engage in expert debates~\cite{xiao2025trading}, and use in-context self-reflection~\cite{koa2024learning}. Some frameworks also apply task-specific fine-tuning~\cite{tian2025tradinggroup, deng2025alphaquanter}. In alpha factor mining, agents utilize iterative refinement~\cite{li2024FAMA, tang2025alphaagent, cao2025chainoflapha} and navigated tree searching~\cite{shi2025navigating} to find effective formulas.

Our method diverges from these asset-centric works by shifting the analytical target from micro-level mapping to macro-level reasoning. While traditional approaches treat assets as independent units and attempt to map noisy features directly to price movements, we posit that individual asset fluctuations are often idiosyncratic. Instead of predicting ``what" an asset will do based on its internal signals, \MYMETHOD~extracts the underlying Investment Arguments that explain ``why" a group of assets moves in tandem, thereby achieving higher reasoning purity by filtering out asset-specific noise.

\subsection{Market-Centric Approaches}

Market-centric approaches examine the financial problem at the stock pool or portfolio level. Traditional research focuses on portfolio construction through inter-stock covariance matrix prediction~\cite{zhu2022forecasting, zheng2023deep} and multi-factor dimension reduction~\cite{lin2021deep}. Some studies apply reinforcement learning for weight optimization across a collection of assets~\cite{zhang2024reinforcement}. Recently, LLM-based systems have been used to synthesize information across the stock pool. For example, some works use diverse expert discussions to analyze constituent stocks for portfolio construction~\cite{tong2024ploutos}. Other methods employ multi-agent collaboration~\cite{guo2025mass}, investor behavior simulation~\cite{yang2025twinmarket}, order-level simulation~\cite{li2025mars} or model-factor collaboration~\cite{li2025rdagentquant} to manage market interactions.

Our framework departs from these market-centric approaches in both perspective and methodology. First, while existing methods are primarily object-oriented as they treat the stock pool as a collection of assets to be weighted, we adopt a logic-oriented perspective that views the market as a dynamic competition of underlying narratives. Second, we shift the focus from spatial allocation to temporal evolution. Instead of merely optimizing asset weights for a static market snapshot, \MYMETHOD~models the complete lifecycle of Modes of Thought. By tracking how investment logics emerge, gain consensus, and decay through temporal alignment, we ensure that portfolio construction is guided by the historical robustness of a narrative rather than transient correlations dependent on specific market states.

\section{Methodology}

\MYMETHOD~is designed to construct daily portfolios from multi-modal financial data. Unlike traditional asset-centric or market-centric methods, \MYMETHOD~is logic-oriented. We focus on the ``why" behind price movements by modeling the market as an evolving ecosystem of investment narratives. The process begins with raw daily data for a universe of $N$ stocks over a period of $T$ trading days. The input for a given stock $s$ on day $t$ is a data point $\mathcal{D}_{t,s}$, comprising information from a set of modalities $M=\{\text{Fundamental}, \text{News}, \text{Technical}\}$:
$$
\mathcal{D}_{t,s} = \{ \mathcal{D}_{t,s,m} \mid m \in M \}
$$
where $\mathcal{D}_{t,s,m}$ is for stock $s$ on day $t$ from modality $m$. The final outputis a daily portfolio weight vector $\mathbf{W}_t \in \mathbb{R}^N$.

\MYMETHOD~operationalizes this process through three sequential stages. First, we extract structured investment arguments from noisy multimodal data. Second, we identify latent modes of reasoning by clustering these arguments and tracking their evolution over time. Third, we evaluate the historical profitability of each mode to construct a final portfolio. The overall architecture of \MYMETHOD~ is provided in Figure~\ref{fig:main}, and the pseudocode is available in Appendix~\ref{appendix::pseudocode_algo}.

\begin{figure*}[t!]
  \centering
  \includegraphics[width=0.79\linewidth]{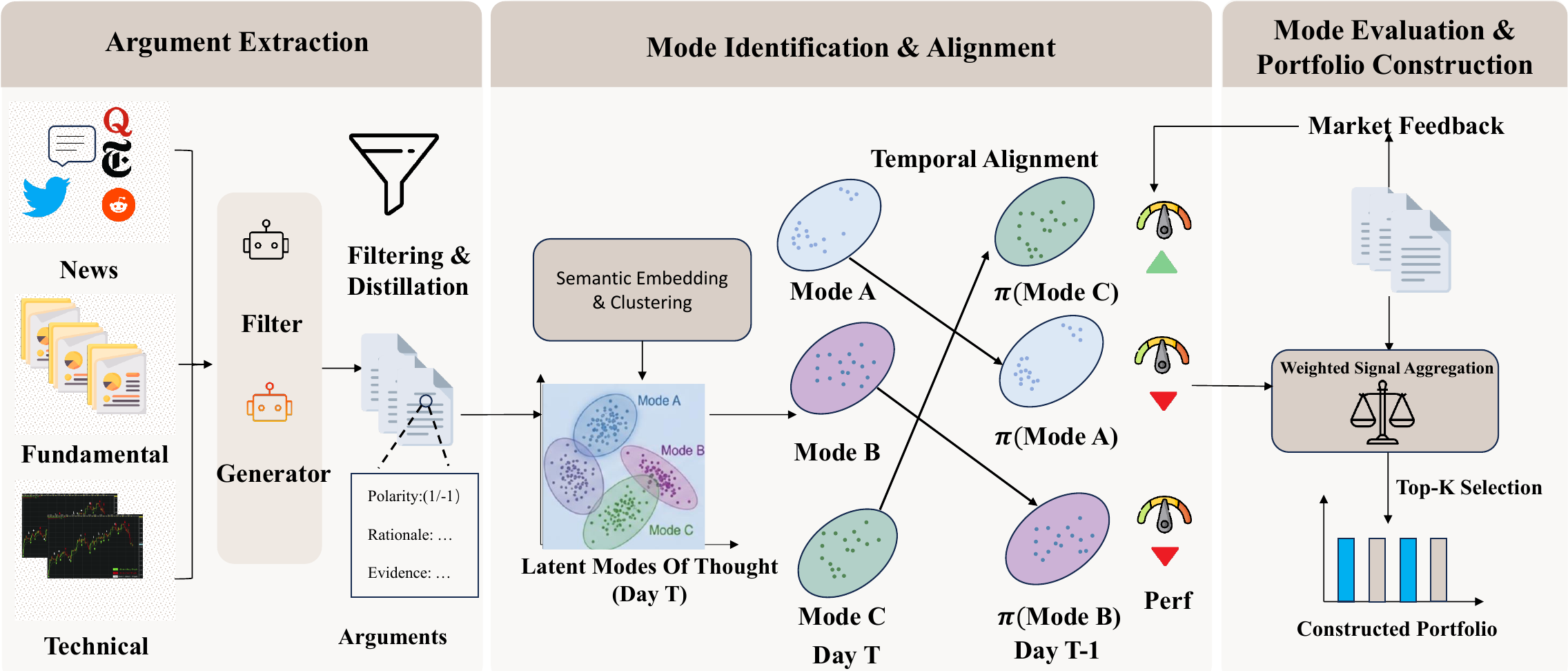}
  \caption{The overall architecture. \MYMETHOD~first mitigates structural noise by distilling raw multimodal data into structured Investment Arguments via multi-agent extraction. To uncover latent consensus, it identifies representative Modes of Thought through GMM-based probabilistic clustering and models market evolution via temporal alignment of these modes across consecutive days. Finally, \MYMETHOD~filters logical noise by evaluating the historical profitability of each mode through an alignment-based scoring mechanism, ensuring that the portfolio is guided by robust and enduring investment narratives.}
  \label{fig:main}
\end{figure*}

\subsection{Argument Extraction: From Noise to Structure}
~\label{method::argument_extraction}

\vspace{-0.5em}
The initial stage of \MYMETHOD~transforms unstructured, multi-modal financial data into structured records, which we term Investment Arguments. This stage is designed to overcome the structural chaos and high noise density in raw data. To manage the low signal-to-noise ratio inherent in raw data, we employ a multi-agent system that filters relevant information from each data modality and then generates structured arguments, each containing a directional stance, a rationale, and supporting evidence. This collaborative design ensures the purity of the reasoning by separating informative signals from irrelevant noise. This process is implemented through two steps: 

\paragraph{Information Filtering.} For each modality $m \in M$, a filter agent, $\mathcal{F}_{\text{filter},m}$, processes the raw data $\mathcal{D}_{t,s,m}$ to extract a concise set of relevant information, $\mathcal{I}_{t,s,m}$.
    \begin{equation}
    \label{equation::filter}
        \mathcal{I}_{t,s,m} = \mathcal{F}_{\text{filter},m}(\mathcal{D}_{t,s,m})
    \end{equation}
\paragraph{Argument Generation.} The filtered information from all modalities is consolidated. A generator agent, $\mathcal{F}_{\text{gen}}$, then synthesizes this unified information set to a series of structured investment arguments, $\{\mathcal{A}_{t,s,i}\}_i$.
\begin{equation}
        \label{equation::generation}
        \{\mathcal{A}_{t,s,i}\}_i = \mathcal{F}_{\text{gen}}\left(\bigcup_{m \in M} \mathcal{I}_{t,s,m}\right)
\end{equation}
Each argument $\mathcal{A}_{t,s,i}$ is a tuple containing a polarity $p$, an analytical rationale $a$, and its supporting evidence $e$:
\begin{equation}
        \label{equation::arguments}
        \mathcal{A}_{t,s,i} = (p_{t,s,i}, a_{t,s,i}, e_{t,s,i})
\end{equation}
where $p_{t,s,i} \in \{+1, -1\}$ indicates the argument's directional stance.The design details are in the Appendix~\ref{appendix::argument}.

\subsection{Mode Identification and Alignment}
\label{method::mode_identification_and_evolution}

After extracting individual arguments, we identify and track the underlying consensus they represent. Isolated arguments are often sparse and may use varied terminology for similar concepts. Besides, the focus of investors may shift over time. Therefore, this stage first identifies latent modes of reasoning by clustering arguments in a semantic space on a given day. Subsequently, it aligns these modes across consecutive days to model their temporal evolution.

\subsubsection{Mode Identification: From Arguments to Latent Consensus}
\label{method::mode_identification}

We model the shared logic among arguments by identifying latent modes within a semantic space. We use a probabilistic approach because market themes often overlap. A single argument might reflect multiple themes simultaneously. To implement this, we apply a Gaussian Mixture Model (GMM) \cite{duda1973pattern, dempster1977maximum} to the collection of daily argument embeddings. This process yields a set of modes, each defined by a Gaussian distribution, and assigns each argument a membership probability to every mode. This captures the nuanced reality of market consensus better than rigid classification.

Formally, we generate a semantic representation for each argument on a given day $t$. We concatenate the argument's rationale $a_{t,s,i}$ and evidence $e_{t,s,i}$, then encode the result into a $D$-dimensional vector $\mathbf{x}_{t,s,i}$ using an encoder $\mathcal{E}$:
\begin{equation}
\label{equation::embedding}
    \mathbf{x}_{t,s,i} = \mathcal{E}(\text{concat}(a_{t,s,i}, e_{t,s,i})) \in \mathbb{R}^D
\end{equation}
We then apply a GMM to the collection of all daily argument embeddings $\{\mathbf{x}_{t,s,i}\}_{s,i}$. This defines $K$ modes for day $t$, where each mode $\mathcal{M}_{t,k}$ is a Gaussian distribution $\mathcal{N}(\boldsymbol{\mu}_{t,k}, \boldsymbol{\Sigma}_{t,k})$. The GMM also computes the posterior probability $\omega_{t,s,i,k}$ that argument $\mathcal{A}_{t,s,i}$ belongs to mode $\mathcal{M}_{t,k}$. The relationship is captured by a vector $\boldsymbol{\omega}_{t,s,i}$:
\begin{equation}
    \label{equation::distribution}
    \boldsymbol{\omega}_{t,s,i} = [\omega_{t,s,i,1}, \dots, \omega_{t,s,i,K}], \quad \text{where} \sum_{k=1}^K \omega_{t,s,i,k} = 1
\end{equation}

\subsubsection{Temporal Mode Alignment: From Snapshot to Evolution}
\label{method::temporal_alignment}

To understand the persistence of themes, we connect the identified modes across time. A static daily analysis is insufficient as the semantic focus of a mode can drift. We address this semantic drift by linking related modes across consecutive days. This allows us to track the complete lifecycle of investment logics as they emerge and decay.

Let $\{\boldsymbol{\mu}_{t-1,k}\}_{k=1}^K$ and $\{\boldsymbol{\mu}_{t,k}\}_{k=1}^K$ be the sets of mode centroids for days $t-1$ and $t$. We seek a permutation $\pi$ that minimizes the total semantic drift between matched modes by applying a linear-assignment algorithm \cite{jonker1987shortest}:
\begin{equation}
    \label{equation::linear_asignment}
    \min_{\pi \in S_K} \sum_{k=1}^{K} d(\boldsymbol{\mu}_{t-1,k}, \boldsymbol{\mu}_{t,\pi(k)})
\end{equation}
where $d(\cdot, \cdot)$ is the Euclidean distance and $S_K$ is the set of all permutations of $\{1, \dots, K\}$. The resulting permutation $\pi$ establishes the temporal correspondence of each mode.

\subsection{Mode Evaluation and Portfolio Construction}
\label{method::mode_evaluation}

The final stage of \MYMETHOD~evaluates the historical effectiveness of each mode by leveraging the mode evolution process and the performance of each mode at every historical cross section. It helps us distinguish between enduring market wisdom and narratives that sound plausible but fail in practice. We develop a mechanism to quantify the historical profitability of each mode. This performance score is then used to weight new signals and generate final stock rankings for the future portfolio.

Formally, at the close of trading on day $t$, we evaluate the performance of all modes from day $t-1$. The one-day forward excess return of stock $s$ is:
\begin{equation}
\label{equation::argument_per}
    r_{t, s} = \frac{c_{t,s} - c_{t-1, s}}{c_{t-1, s}} - \frac{1}{N}\sum_{i=1}^{N} \frac{c_{t,i} - c_{t-1, i}}{c_{t-1, i}}
\end{equation}
where $c_{t,s}$ is the closing price at day $t$. The realized score of an argument $\mathcal{A}_{t-1,s,i}$ is:
\begin{equation}
\label{equation::score_calc}
    \text{Score}(\mathcal{A}_{t-1,s,i}) = p_{t-1,s,i} \cdot r_{t,s}
\end{equation}
The daily aggregated score for each mode $k$ on day $t-1$, $\text{AggScore}_{t-1,k}$, is the responsibility-weighted average score of all arguments:
\begin{equation}
\label{equation::aggscore_calc}
    \text{AggScore}_{t-1,k} = \frac{\sum_{s,i} \omega_{t-1,s,i,k} \cdot \text{Score}(\mathcal{A}_{t-1,s,i})}{\sum_{s,i} \omega_{t-1,s,i,k}}
\end{equation}
To filter out fleeting, high-frequency anomalies, we update a long-term performance score, $\text{Perf}_{t-1,k}$, using an exponential moving average. This update uses the temporal mapping $\pi'$ from the prior step, which links day $t-2$ to $t-1$, to ensure performance continuity:
\begin{equation}
\label{eq:perf_update}
    \text{Perf}_{t-1,k} = \lambda \cdot \text{Perf}_{t-2, \pi'^{-1}(k)} + (1-\lambda) \cdot \text{AggScore}_{t-1, k}
\end{equation}
where $\lambda \in [0, 1]$ is a smoothing factor.

Finally, we construct our daily portfolio $\mathbf{W}_t$. By using historically proven narratives to weight new signals, we ensure that our decisions are guided by robust reasoning. The process begins by quantifying the alignment of each new argument $\mathcal{A}_{t,s,i}$ with the established modes from day $t-1$. This alignment is computed as a posterior probability, $\hat{\omega}_{t,s,i,k}$, using the GMM parameters $\{\pi_{t-1,k}, \boldsymbol{\mu}_{t-1,k}, \boldsymbol{\Sigma}_{t-1,k}\}_{k=1}^K$ from the previous day:
\begin{equation}
    \label{equation::distribution_calc}
    \hat{\omega}_{t,s,i,k} = \frac{\pi_{t-1,k} \mathcal{N}(\mathbf{x}_{t,s,i} | \boldsymbol{\mu}_{t-1,k}, \boldsymbol{\Sigma}_{t-1,k})}{\sum_{j=1}^{K} \pi_{t-1,j} \mathcal{N}(\mathbf{x}_{t,s,i} | \boldsymbol{\mu}_{t-1,j}, \boldsymbol{\Sigma}_{t-1,j})}
\end{equation}
This probability distribution then weights the long-term performance of the modes, $\text{Perf}_{t-1,k}$, to produce a predicted score for the new argument:
\begin{equation}
    \label{equation::Score_calc}
    \widehat{\text{Score}}(\mathcal{A}_{t,s,i}) = \sum_{k=1}^{K} \hat{\omega}_{t,s,i,k} \cdot \text{Perf}_{t-1, k}
\end{equation}
We then aggregate these argument-level scores to compute a composite signal for each stock $s$. Let $\mathcal{I}_{t,s}^+ = \{i \mid p_{t,s,i} = +1\}$ and $\mathcal{I}_{t,s}^- = \{i \mid p_{t,s,i} = -1\}$ be the index sets for positive and negative arguments. The signal is the difference between the average predicted scores of its positive and negative arguments:
\begin{equation}
\label{equation::signal}
\begin{aligned}
    \text{Signal}_{t,s} = & \frac{1}{|\mathcal{I}_{t,s}^+| + \epsilon} \sum_{i \in \mathcal{I}_{t,s}^+} \widehat{\text{Score}}(\mathcal{A}_{t,s,i}) \\
    & - \frac{1}{|\mathcal{I}_{t,s}^-| + \epsilon} \sum_{i \in \mathcal{I}_{t,s}^-} \widehat{\text{Score}}(\mathcal{A}_{t,s,i})
\end{aligned}
\end{equation}
where $\epsilon$ is a small constant for numerical stability. The resulting signal vector, $\text{Signal}_{t} \in \mathbb{R}^N$, provides a ranking for all stocks. Based on this ranking, we employ a top-K strategy with equal weight to construct the portfolio $ \mathbf{W}_t$ at the open of the next trading day.

\section{Experiments}

\subsection{Setup}

\begin{table*}[t!]
    \centering
    \small 
    \caption{Performance comparison on the SSE 50, CSI 300, and CSI 500 stock pools. \textbf{Bold} and \underline{underlined} numbers represent the best and second-best performance, respectively. The symbols $^{\uparrow}$ and $^{\downarrow}$ indicate that higher and lower values are better, respectively. The \textit{Benchmark Index} represents the buy-and-hold strategy of the corresponding market index.}
    \label{tab:performance_comparison}
    \setlength{\tabcolsep}{8pt} 
    \renewcommand{\arraystretch}{0.99} 
    \begin{tabular}{@{}llccccccc@{}} 
        \toprule
        \multirow{2}{*}{\textbf{Dataset}} & \multirow{2}{*}{\textbf{Method}} & \multicolumn{4}{c}{\textbf{Correlation Metrics (\%)}} & \multicolumn{3}{c}{\textbf{Portfolio Metrics}} \\
        \cmidrule(lr){3-6} \cmidrule(lr){7-9}
        & & \textit{IC}\rlap{$^{\uparrow}$} & \textit{ICIR}\rlap{$^{\uparrow}$} & \textit{RIC}\rlap{$^{\uparrow}$} & \textit{RICIR}\rlap{$^{\uparrow}$} & \textit{AR}\rlap{$^{\uparrow}$} & \textit{MDD}\rlap{$^{\downarrow}$} & \textit{SR}\rlap{$^{\uparrow}$} \\
        \midrule
        
        \multirow{9}{*}{SSE 50} 
        & LightGBM       & 2.63 & 13.56 & 2.98 & 17.14 & 21.19\% & 15.26\% & 0.9779 \\
        & DTML           & 2.97 & 13.89 & 3.19 & 14.53 & 18.47\% & 16.17\% & 0.9767 \\
        & FactorVAE      & 3.05 & 13.25 & 3.56 & 13.98 & 19.06\% & 16.55\% & 0.9263 \\
        & Master         & 4.30 & 18.42 & \underline{5.42} & 24.84 & 24.64\% & \textbf{10.71}\% & \underline{1.3761} \\
        & SEP            & 3.65 & 17.89 & 3.16 & 15.30 & 22.58\% & 16.62\% & 0.9812\\
        & TradingAgents  & \underline{4.32} & \underline{21.50} & 5.31 & 25.27 & \underline{24.78}\% & 11.05\% & 1.3543 \\
        & R\&D-Agent(Q)  & 4.27 & 20.89 & 5.03 & \underline{25.80} & 22.70\% & 11.21\% & 1.2988 \\
        \rowcolor{gray!15} 
        & \textbf{\MYMETHOD} & \textbf{6.31} & \textbf{32.99} & \textbf{5.48} & \textbf{29.93} & \textbf{28.74\%} & \underline{10.89\%} & \textbf{1.4021} \\
        & \textit{Benchmark Index} & \multicolumn{1}{c}{\textendash} & \multicolumn{1}{c}{\textendash} & \multicolumn{1}{c}{\textendash} & \multicolumn{1}{c}{\textendash} & 16.18\% & 13.95\% & 0.8601 \\
        \midrule
        
        \multirow{9}{*}{CSI 300} 
        & LightGBM       & 1.18 & 9.86 & 2.15 & 17.50 & 12.89\% & 19.08\% & 0.5864 \\
        & DTML           & 3.51 & 23.25 & 4.16 & 25.58 & 19.13\% & 15.61\% & 0.8788 \\
        & FactorVAE      & 3.47 & 20.81 & 3.98 & 26.82 & 18.79\% & 16.20\% & 0.7892 \\
        & Master         & 4.37 & 22.60 & 5.26 & \underline{28.30} & 24.46\% & 19.71\% & 1.0891 \\
        & SEP            & 2.98 & 19.76 & 3.42 & 24.75 & 18.22\% & 16.98\% & 0.9730 \\
        & TradingAgents  & 4.18 & 23.67 & 5.09 & 25.78 & 23.69\% & 18.40\% & 1.1310 \\
        & R\&D-Agent(Q)  & \underline{4.85} & \underline{28.72} & \underline{5.45} & 27.81 & \underline{25.02}\% & \underline{16.50}\% & \underline{1.3216} \\
        \rowcolor{gray!15}
        & \textbf{\MYMETHOD} & \textbf{6.80} & \textbf{41.71} & \textbf{6.16} & \textbf{39.75} & \textbf{32.48\%} & \textbf{10.76\%} & \textbf{1.5161} \\
        & \textit{Benchmark Index} & \multicolumn{1}{c}{\textendash} & \multicolumn{1}{c}{\textendash} & \multicolumn{1}{c}{\textendash} & \multicolumn{1}{c}{\textendash} & 15.72\% & 17.74\% & 0.7265 \\
        \midrule
        
        \multirow{9}{*}{CSI 500} 
        & LightGBM       & 2.18 & 14.01 & 4.81 & 33.11 & 17.41\% & 24.52\% & 0.6872 \\
        & DTML           & 3.64 & 30.47 & 4.18 & \underline{34.72} & 21.04\% & 23.27\% & 0.7971 \\
        & FactorVAE      & 2.98 & 26.25 & 4.97 & 34.63 & 20.72\% & 24.99\% & 0.7104 \\
        & Master         & 2.74 & 15.63 & 4.60 & 23.21 & 16.36\% & 26.64\% & 0.6570 \\
        & SEP            & 2.88 & 18.76 & 3.95 & 21.86 & 15.78\% & 25.32\% & 0.5865\\
        & TradingAgents  & 4.06 & \underline{31.56} & \underline{5.01} & 33.06 & 23.29\% & \underline{18.64}\% & \underline{1.0517} \\
        & R\&D-Agent(Q)  & \underline{4.15} & 29.09 & 4.88 & 32.19 & \underline{23.50}\% & 19.89\% & 1.0343 \\
        \rowcolor{gray!15}
        & \textbf{\MYMETHOD} & \textbf{6.72} & \textbf{41.15} & \textbf{6.64} & \textbf{34.75} & \textbf{31.97\%} & \textbf{16.18\%} & \textbf{1.2326} \\
        & \textit{Benchmark Index} & \multicolumn{1}{c}{\textendash} & \multicolumn{1}{c}{\textendash} & \multicolumn{1}{c}{\textendash} & \multicolumn{1}{c}{\textendash} & 17.90\% & 24.55\% & 0.6824 \\
\bottomrule
\end{tabular}
\end{table*}

\textbf{Datasets}: We evaluate \MYMETHOD~on three A-share stock universes with varying styles: the SSE 50, CSI 300, and CSI 500. Our experiments run from 2023 Q4 to 2025 Q4, a period that encompasses multiple market style transitions, including the bear market of late 2023 to early 2024, a comprehensive bull market in September 2024, a structured market phase in mid-2025, and an out-of-LLM-knowledge period in 2025 Q4. The details of the dataset description and  composition are provided in the Appendix~\ref{appendix::dataset}.

\textbf{Performance metrics}: Following previous work~\cite{yoo2021accurate,li2024master,guo2025mass, li2025rdagentquant}, we utilize two types of metrics to evaluate the performance of~\MYMETHOD: (1) Correlation-based metrics: Information Coefficient (IC), IC Information Ratio (ICIR), Rank Information Coefficient (RIC), and RIC Information Ratio (RICIR).  For these correlation metrics, the prediction target is daily forward return. (2) Portfolio-based metrics: Annualized Return (AR), Maximum Drawdown (MDD), and Sharpe Ratio (SR). The details of these metrics are in the Appendix~\ref{appendix::metrics}.

\textbf{Baselines}: We compare \MYMETHOD~with various competitive baselines across different types: (1) Traditional machine learning methods: LightGBM~\cite{lgbm}. (2) Deep learning techniques: DTML~\cite{yoo2021accurate}, FactorVAE~\cite{duan2022factorvae}, MASTER~\cite{li2024master}. (3) Agent-based approaches centered on a single asset: SEP~\cite{koa2024learning}, TradingAgents~\cite{xiao2025tradingagents}. (4) Agent-based approaches centered on the market: R\&D-Agent(Q)~\cite{li2025rdagentquant}. The details of all baselines are provided in the Appendix~\ref{appendix::baseline}.

\textbf{Implementation Details}: DeepSeek V3.1~\cite{deepseekai2024deepseekv3technicalreport} serves as the backbone LLM in \MYMETHOD~and all agent-based baselines for fair comparisons. We use Qwen3 Embedding-8B~\cite{qwen3embedding} as the embedding model in mode identification. The number of modes $K$ is set to 15 for the SSE 50 and 20 for the CSI 300 and the CSI 500. The smoothing factor $\lambda$ is set to 0.5. The small constant $\epsilon$ is set to 1e-5. For backtest experiments, we utilize an index-enhanced strategy. The transaction cost is set to 1.5e-4, and we update our portfolio daily by holding the top 20\% stocks in Equation~\ref{equation::signal}.

\begin{figure}[!htbp]
    \centering
    \vspace{-1em}
    \includegraphics[width=0.86\linewidth]{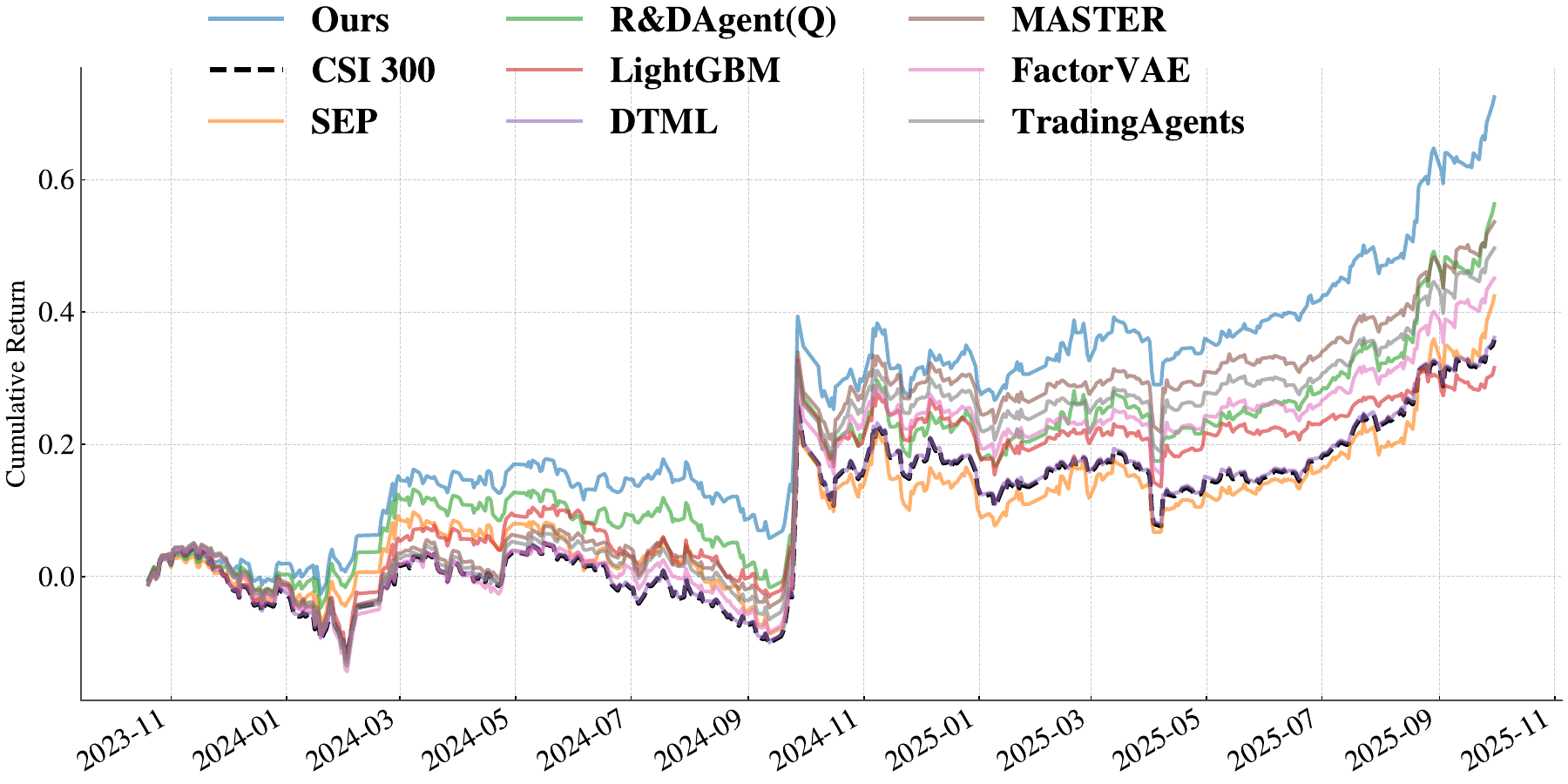}
    \caption{Backtest on the CSI 300. We compare \MYMETHOD~ with all baselines and the CSI 300 Index.}
    \vspace{-1em}
    \label{fig:csi_300_backtest}
\end{figure}

\subsection{Main Experiments}

We conduct the main experiments on three datasets from 2023 Q4 to 2025 Q3. Table~\ref{tab:performance_comparison} exhibits all experiment results across three different datasets. Our analysis yields two key findings. First, \MYMETHOD~ demonstrates strong predictive capabilities for stock returns, consistently achieving higher IC, RIC, and AR compared to all baselines. Second, \MYMETHOD~shows improved stability across different market conditions, obtaining higher ICIR, RICIR, and SR while also reducing the MDD.

Besides, we also draw the complete backtest curve on the CSI 300. As shown in Figure~\ref{fig:csi_300_backtest}, \MYMETHOD~takes the performance lead in most backtest periods. Furthermore, \MYMETHOD~demonstrates enhanced robustness by sustaining low drawdowns during the early 2024 liquidity crisis and the April 2025 trade frictions, while also achieving favorable returns throughout the bull market in the mid-2025.

\subsection{Experiments on Data Leakage Concern}

To validate \MYMETHOD~against potential data leakage, we conduct a temporal out-of-sample test on three datasets using new data from Q4 2025. The results in Table~\ref{tab:performance_comparison_2025q4} demonstrate that \MYMETHOD~maintains its performance advantage over all baselines. This strongly indicates that \MYMETHOD~effectiveness stems from our methodological design and not from contamination within the LLM's training corpus.
\begin{table*}[!h]
    \centering
    \small 
    \caption{Experiments on 2025 Q4. Performance comparison on three datasets after the LLM knowledge cutoff. }
    \label{tab:performance_comparison_2025q4}
    \setlength{\tabcolsep}{8pt} 
    \renewcommand{\arraystretch}{1} 
    \begin{tabular}{@{}llccccccc@{}} 
        \toprule
        \multirow{2}{*}{\textbf{Dataset}} & \multirow{2}{*}{\textbf{Method}} & \multicolumn{4}{c}{\textbf{Correlation Metrics (\%)}} & \multicolumn{3}{c}{\textbf{Portfolio Metrics}} \\
        \cmidrule(lr){3-6} \cmidrule(lr){7-9}
        & & \textit{IC}\rlap{$^{\uparrow}$} & \textit{ICIR}\rlap{$^{\uparrow}$} & \textit{RIC}\rlap{$^{\uparrow}$} & \textit{RICIR}\rlap{$^{\uparrow}$} & \textit{AR}\rlap{$^{\uparrow}$} & \textit{MDD}\rlap{$^{\downarrow}$} & \textit{SR}\rlap{$^{\uparrow}$} \\
        \midrule
        
        \multirow{9}{*}{SSE 50} 
        & LightGBM       & 2.26 & 20.33 & 2.37 & 22.24 & \underline{8.10\%} & 4.89\% & \underline{0.8523} \\
        & DTML           & 2.45 & 15.81 & 3.24 & \underline{22.91} & 5.80\% & 5.53\% &  0.5334\\
        & FactorVAE      & 1.79 & 10.60 & \textbf{2.88} & 19.85 & 2.84\% & 5.28\% & 0.4998 \\
        & Master         & 2.37 & \textbf{27.96} & 2.48 & 22.08 & 6.49\% & \underline{4.81\%} & 0.6820 \\
        & SEP            & 2.30 & 18.55 & 1.91 & 17.30 & 3.03\% & 6.06\% &  0.4120\\
        & TradingAgents  &  2.18 &  20.06 & 2.05 & 18.87 & 4.18\% & 5.64\% &  0.5226\\
        & R\&D-Agent(Q)  & \underline{2.58} & 23.88 & 2.65 &  21.80 & 7.88\% & 6.10\% & 0.6851 \\
        \rowcolor{gray!15} 
        & \textbf{\MYMETHOD} & \textbf{2.67} & \underline{24.35} & \underline{2.83} & \textbf{26.42} & \textbf{8.25\%} & \textbf{4.79\%} & \textbf{0.8894} \\
        & \textit{Benchmark Index} & \multicolumn{1}{c}{\textendash} & \multicolumn{1}{c}{\textendash} & \multicolumn{1}{c}{\textendash} & \multicolumn{1}{c}{\textendash} & 5.66\%  & 4.85\%  & 0.7653 \\
        \midrule

        \multirow{9}{*}{CSI 300} 
        & LightGBM       & 1.16 & 9.84 & 3.02 & 20.54 & -2.56\% & 7.42\% & -0.2579 \\
        & DTML           & 3.44 & 21.08 & 4.27 & \underline{26.47} & 3.20\% & \underline{5.01\%} & 0.4789 \\
        & FactorVAE      & 2.28 & 16.62 & 3.70 & 21.22 & 0.99\% & 5.89\% & 0.1335 \\
        & Master         & 3.47 & 17.18 & 3.77 & 18.13 & 0.79\% & 6.16\% & 0.0924 \\
        & SEP            & 1.93 & 11.72 & 2.34 & 14.29 & 2.06\% & 5.43\% & 0.1910 \\
        & TradingAgents  & 2.61 & 15.34 & 3.07 & 19.89 & 4.87\% & 5.12\% & \underline{0.4120}\\
        & R\&D-Agent(Q)  & \underline{4.09} & \underline{23.01} & \underline{3.89} & 26.20 & \underline{5.42\%} & 5.31\% & 0.5838 \\
        \rowcolor{gray!15} 
        & \textbf{\MYMETHOD} & \textbf{4.12 }& \textbf{32.11} & \textbf{4.53} &\textbf{ 29.39} & \textbf{9.77\%} &\textbf{ 4.87\% }&\textbf{ 0.7353} \\
        & \textit{Benchmark Index} & \multicolumn{1}{c}{\textendash} & \multicolumn{1}{c}{\textendash} & \multicolumn{1}{c}{\textendash} & \multicolumn{1}{c}{\textendash} & -0.92\% & 7.46\% & -0.0837\\
        \midrule
        \multirow{9}{*}{CSI 500} 
        & LightGBM       & 2.39 & 12.24 & 4.71 & 28.40 & 2.39\% & 8.73\% & 0.0618 \\
        & DTML           & 3.24 & 20.49 & 1.89 & 16.54 & 14.06\% & 9.02\% & 0.6495 \\
        & FactorVAE      & 2.23 & 18.80 & 3.66 & 17.75 & 12.51\% & 8.75\% & 0.6122 \\
        & Master         & 2.88 & 20.42 & 4.65 & 33.28 & \underline{19.84\%} & \underline{6.21\%} & \underline{1.1254} \\
        & SEP            & 3.01 & 16.35 & 3.24 & 19.49 & 4.71\% & 8.96\% & 0.2427 \\
        & TradingAgents  & 3.87 & 21.64 & 3.93 & 28.32 & 8.22\% & 8.40\% & 0.5680  \\
        & R\&D-Agent(Q)  & \underline{4.38} & \underline{27.70} & \underline{4.73} & \underline{33.59} & 13.98\% & 7.22\% & 0.9606  \\
        \rowcolor{gray!15} 
        & \textbf{\MYMETHOD} & \textbf{5.15} & \textbf{32.61 }& \textbf{5.31} & \textbf{34.97} & \textbf{20.56\%} & \textbf{5.95\% }& \textbf{1.1433} \\
        & \textit{Benchmark Index} & \multicolumn{1}{c}{\textendash} & \multicolumn{1}{c}{\textendash} & \multicolumn{1}{c}{\textendash} & \multicolumn{1}{c}{\textendash} & 2.88\%  &  10.08\% & 0.2281\\
\bottomrule
\end{tabular}
\end{table*}

\subsection{Ablation Study}
We conduct an ablation study to evaluate the contribution of each core component of ~\MYMETHOD. Table~\ref{tab:ablation_components} presents the results of incrementally adding each component: Structured Argument Generation (SAG), Modes of Thought (MOT), Probabilistic Modeling (PM), and Temporal Alignment (TA).

The first row establishes a baseline by using the LLM directly on raw data, which yields the lowest performance across all datasets and demonstrates the necessity of structured analysis. Introducing SAG alone provides a performance lift on smaller stock pools like SSE 50. However, its effectiveness diminishes on the larger CSI 500, where metrics show minimal improvement, suggesting that simply extracting arguments is insufficient for complex scenarios.

The integration of MOT, even with a deterministic clustering approach, leads to a substantial performance leap, particularly on the CSI 300 and CSI 500 datasets. This highlights the critical role of organizing arguments into coherent reasoning patterns. Subsequently, replacing the deterministic clustering with our Probabilistic Modeling component further enhances performance across all datasets. This confirms the benefit of modeling the ambiguous relationship between arguments and their underlying modes of thought.

Finally, adding the Temporal Alignment to form the complete model also brings significant gains. For instance, on SSE 50, the IC nearly doubles from 3.18 to 6.31, and the ICIR doubles from 16.95 to 32.99. This result validates the hypothesis that modeling the lifecycle of thought patterns is essential for generating robust predictive signals.

\newcommand{\cmark}{\ding{51}}
\newcommand{\xmark}{\ding{55}}
\begin{table*}[t]
\centering
\caption{Ablation study of different components on three datasets. We incrementally add components SAG (Structured Argument Generation), MOT (Mode Of Thoughts), PM (Probabilistic Modeling), and TA (Temporal Alignment) to evaluate their contribution to the performance metrics (IC, ICIR, RIC, and RICIR). All Values are in percent.}
\label{tab:ablation_components}
\resizebox{\textwidth}{!}{%
\renewcommand{\arraystretch}{1} 
\begin{tabular}{cccc|cccccccccccc}
\toprule
\multicolumn{4}{c}{\textbf{Components}} & \multicolumn{4}{c}{\textbf{SSE 50}} & \multicolumn{4}{c}{\textbf{CSI 300}} & \multicolumn{4}{c}{\textbf{CSI 500}} \\
\cmidrule(lr){1-4} \cmidrule(lr){5-8} \cmidrule(lr){9-12} \cmidrule(lr){13-16}
SAG & MOT & PM & TA & IC & ICIR & RIC & RICIR & IC & ICIR & RIC & RICIR & IC & ICIR & RIC & RICIR \\
\midrule
\xmark & \xmark & \xmark & \xmark & 1.87 & 10.11  &1.06 & 7.89  & 0.79 & 6.92 & 0.94 & 5.88 & 0.18 & 1.68 & 0.24 & 3.63 \\
\cmark & \xmark & \xmark & \xmark & 2.51 & 12.73 & 1.92 & 9.33 & 3.16 & 14.40 & 1.89 & 8.65 & 0.21 & 1.65 & 0.38 & 4.70 \\
\cmark & \cmark & \xmark & \xmark & 2.47 & 14.66 & 1.91 & 10.85  & 5.30 & 30.43 & 5.24 & 29.36 & 4.90 & 38.09 & 4.87 & 36.20 \\
\cmark & \cmark & \cmark & \xmark & 3.18 & 16.95 & 3.08 & 17.38  & 6.44 & 38.59 & 5.61 &32.67 & 5.22  & 37.14 & 5.10 & 34.12 \\
\rowcolor{gray!20} 
\cmark & \cmark & \cmark & \cmark & \textbf{6.31} & \textbf{32.99} & \textbf{5.48}& \textbf{29.93} & \textbf{6.80} & \textbf{41.71}  &\textbf{ 6.16} & \textbf{39.75 }& \textbf{6.72} &  \textbf{41.15} & \textbf{6.64} & \textbf{34.75} \\
\bottomrule
\end{tabular}%
}
\end{table*}

\subsection{Parameter Sensitivity Analysis}
\begin{figure}[htbp!]
    \centering
    \includegraphics[width=0.75\linewidth]{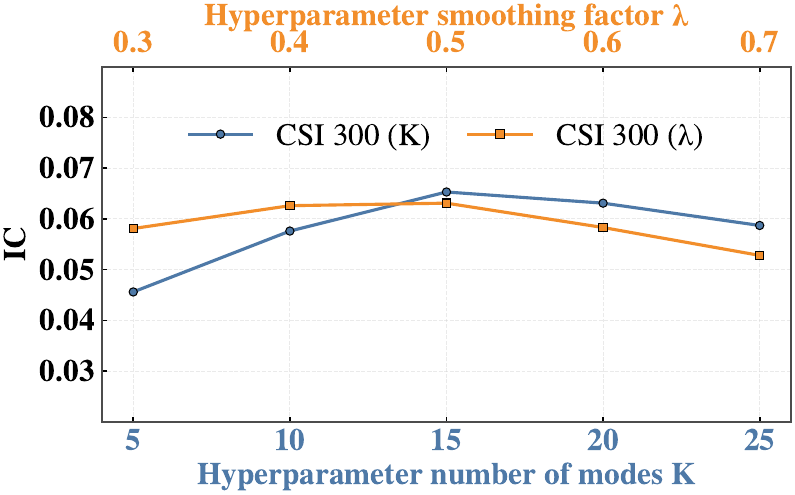}
    \caption{\MYMETHOD~exhibits a moderate sensitivity.}
    \vspace{-1em}
    \label{fig:ParameterSensitivityExperiments_ic}
\end{figure}

We investigate the sensitivity of \MYMETHOD~to its key hyperparameters: the number of modes $K$ and the smoothing factor $\lambda$. The analysis is conducted on the CSI 300, and the results are presented in Figure~\ref{fig:ParameterSensitivityExperiments_ic}.

Overall, \MYMETHOD~demonstrates robustness to variations in both parameters, maintaining stable performance across a considerable range of values. 

For the number of modes $K$, performance degrades when the value is too small. An insufficient $K$ forces conceptually distinct arguments into the same mode, which negatively impacts the quality of the resulting signals. Regarding the smoothing factor $\lambda$, excessively large values also harm performance. As defined in Equation~\ref{eq:perf_update}, a large $\lambda$ places greater weight on the long-term historical performance of modes relative to their most recent effectiveness. This reduces the model's responsiveness to recent shifts in the predictive power of the thought patterns, thereby degrading the final results. More results are provided in Figure~\ref{fig:ParameterSensitivityExperiments_ric}.
\subsection{Cost Time Analysis}
We compare the daily average running time of \MYMETHOD~with SOTA LLM-based baselines. As shown in Figure~\ref{fig:runtime} in the Appendix, \MYMETHOD~achieves lower computational cost by filtering noise via structured argument extraction and by aligning information across assets and trading days through mode identification and alignment, thereby reducing the overhead of explicit LLM reflection.
\subsection{Case Study}

\begin{figure}[htbp]
    \centering
    \includegraphics[width=0.84\linewidth]{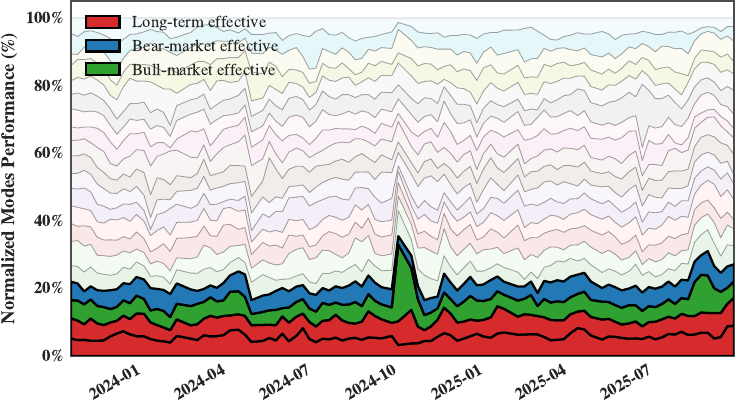}
    \caption{Temporal evolution of normalized performance for different categories of Modes of Thought on the CSI 300. The stack plot illustrates the dynamic lifecycle and shifting prevalence of modes across different market conditions.}
    \vspace{-1em}
    \label{fig:mode_visualization}
\end{figure}

To investigate the lifecycle and nature of the identified Modes of Thought, we conduct a case study on the CSI 300 dataset. We track the long-term performance score, $\text{Perf}$, for each mode as defined in Equation~\ref{eq:perf_update}. A mode is considered ``effective" on a given day if its $\text{Perf}$ is positive, signifying that it is capturing excess returns according to Equation~\ref{equation::argument_per}, ~\ref{equation::score_calc} and ~\ref{equation::aggscore_calc}. Based on their sustained effectiveness across distinct market regimes—a bear market (Oct 2023 - Sep 2024) and a bull market (Sep 2024 - Present)—we classify modes into four categories: 1) Long-term effective, 2) Bull-market effective, 3) Bear-market effective, and 4) Ineffective. In Appendix~\ref{appendix::modes_details}, we give the detailed classification methodology of the above four categories.

Figure~\ref{fig:mode_visualization} presents a temporal analysis of these categories. The x-axis represents the trading days spanning the entire study period. The y-axis displays the daily aggregated and normalized performance share of each category. This is constructed by applying a Softmax function to the $\text{Perf}$ scores of all modes for a given day. Consequently, the total height of the stacked plot at any point in time sums to one, with the area of each colored band representing the relative performance contribution of its corresponding category to the collective performance of all active modes.

The visualization reveals several key dynamics. First, a large portion of modes remain ineffective, indicating that most emergent investment logics do not consistently generate alpha. This aligns with the observation that sustained profitability in financial markets is rare. Second, the composition of effective modes evolves with market conditions. Modes effective in the bear market proved resilient during the early 2024 liquidity crisis. Conversely, modes effective in the bull market, which correspond to trend-following and sentiment-driven logic, gained dominance during the market rally in late 2024. The former's underperformance in the bull market highlights its inability to capture sentiment-driven premiums. This case study demonstrates \MYMETHOD's capacity to identify and track the evolving efficacy of distinct investment narratives. In Appendix~\ref{appendix::modes_details}, we give detailed examples of three categories of effective modes.

\section{Conclusion}
In this paper, we introduce~\MYMETHOD, a LLM-based framework for financial trading. \MYMETHOD~leverages a new Strategy-Centric viewpoint by mining, tracking, and evaluating the lifecycle of different evolving investment logics. Our approach gains enhanced performance and stability against competitive baselines and shows promising robustness and generalizability under different market regimes.


\nocite{langley00}

\bibliography{main}
\bibliographystyle{icml2025}

\newpage
\appendix
\newpage
\onecolumn

\section{Appendix}
\subsection{Pseudocode of \MYMETHOD}
\label{appendix::pseudocode_section}

\begin{algorithm}[H]
\caption{\MYMETHOD}\label{appendix::pseudocode_algo}
\KwIn{Raw multi-modal data $\{\mathcal{D}_t\}_{t \in T}$, historical prices $\{P_t\}_{t \in T}$, number of modes $K$, EMA factor $\lambda$}
\KwOut{Sequence of daily portfolios $\{\mathbf{W}_t\}_{t \in T}$}

Initialize long-term performance scores $\text{Perf}_{0,k} \leftarrow 0$ for $k=1, \dots, K$\;
Initialize mode parameters for day 0, $\text{Modes}_0 \leftarrow \emptyset$\;

\For{each trading day $t \in T$}{
  \tcp{--- Stage 1: Argument Extraction ---}
  Extract structured arguments $\{\mathcal{A}_{t,s,i}\}_{s,i}$ from raw data $\mathcal{D}_t$ via Eqs. (\ref{equation::filter}-\ref{equation::arguments})\;
  
  \tcp{--- Stage 2: Mode Identification and Alignment ---}
  Encode arguments to get embeddings $\{\mathbf{x}_{t,s,i}\}$ via Eq. (\ref{equation::embedding})\;
  
  Fit GMM on $\{\mathbf{x}_{t,s,i}\}$ to identify daily modes $\text{Modes}_t = \{\mathcal{M}_{t,k}\}_{k=1}^K$ and responsibilities $\{\boldsymbol{\omega}_{t,s,i}\}$\;
  
  \If{$t > 0$}{
    \tcp{Align today's modes with yesterday's}
    Find permutation $\pi_t$ by aligning centroids of $\text{Modes}_t$ and $\text{Modes}_{t-1}$ via Eq. (\ref{equation::linear_asignment})\;

    \tcp{--- Stage 3.1: Mode Evaluation (Backward-looking) ---}
    \tcp{Update performance scores using t-1's predictions and t's returns}
    Calculate aggregated score $\text{AggScore}_{t-1,k}$ for each mode from day $t-1$ via Eqs. (\ref{equation::argument_per}-\ref{equation::aggscore_calc})\;
    Update long-term performance $\text{Perf}_{t-1,k}$ using $\text{Perf}_{t-2}$ and $\pi_{t-1}$ via EMA in Eq. (\ref{eq:perf_update})\;
    
    \tcp{--- Stage 3.2: Portfolio Construction (Forward-looking) ---}
    \tcp{Generate trading signals for day t using updated scores}
    Calculate predicted score $\widehat{\text{Score}}(\mathcal{A}_{t,s,i})$ for each new argument via Eqs. (\ref{equation::distribution_calc}-\ref{equation::Score_calc}) using $\text{Perf}_{t-1}$\;
    Aggregate argument scores into a final stock signal $\text{Signal}_{t,s}$ via Eq. (\ref{equation::signal})\;
    Construct portfolio $\mathbf{W}_t$ based on a Top-K ranking of $\text{Signal}_{t}$\;
  }
  \Else{
    $\mathbf{W}_t \leftarrow \text{EqualWeighted Portfolio}$
  }
}
\Return{Sequence of daily portfolios $\{\mathbf{W}_t\}_{t \in T}$}\;
\end{algorithm}

\subsection{Implementation details of Argument Extraction}
\label{appendix::argument}
\subsubsection{Implementation details of Filter Agent}
\begin{tcolorbox}[
  enhanced,                   
  breakable,                  
  colback=EFE9E3,             
  colframe=C9B59C,     
  boxrule=1.0pt,              
  arc=2mm,                    
  left=4pt, right=4pt,        
  top=4pt, bottom=4pt,
  width=\linewidth,           
  title={\bfseries System \& User Prompts},
  fonttitle=\bfseries
]
{\ttfamily\small\justifying
\textbf{System prompts}

You are an expert on financial market analysis, portfolio management and quantitative trading. Strictly follow the user's requirements to generate sound, logical and detailed responses. Your ouput should only contain a JSON object.
                                Do not include any explanations, addditional text, notes, code block markers like ``` or ```json.

\textbf{User prompts for Information Filtering.}

\textbf{Context}: You are part of an automated investment analysis framework. Your analysis will serve as a primary input for a downstream synthesis agent. The goal is to reduce noise and provide a clear, factual summary of the current situation for \textbf{[Asset Name] ([Asset Ticker])} as of\textbf{[Analysis Date]}.

\textbf{Input Data}:

You will be provided with raw [Modality Name] data.
[Raw Data for the specified modality, e.g., a table of fundamental ratios, a list of recent news headlines, or a time-series of technical indicators]

\textbf{Task}:
1.  \textbf{Analyze}: Thoroughly review the provided raw data.

2.  \textbf{Identify Salience}: Identify the most salient points, patterns, and anomalies that could materially impact the asset's future price performance. Disregard trivial or redundant information.

3. \textbf{ Summarize}: Distill your findings into a concise, structured summary. The summary must be objective and based strictly on the provided data.

\textbf{Constraints}:

1. Focus exclusively on the provided data. Do not introduce external information or make speculative predictions.

2. The output must be a factual distillation, not a final investment recommendation.

3. Present the information in a clear, structured format using the specified headings.

\textbf{Output Format}:

\{{

"Modality\_name": [Modality Name],

"Analysis\_summary": "...",

"Asset\_code": [Asset Ticker]

\}}

}

\end{tcolorbox}

\subsubsection{Implementation details of Generator Agent}

\begin{tcolorbox}[
  enhanced,                   
  breakable,                  
  colback=EFE9E3,             
  colframe=C9B59C,     
  boxrule=1.0pt,                
  arc=2mm,                    
  left=4pt, right=4pt,        
  top=4pt, bottom=4pt,
  width=\linewidth,           
  title={\bfseries System \& User Prompts},
  fonttitle=\bfseries
]
{\ttfamily\small\justifying

\textbf{User prompts for Argument Generation.}

\textbf{Context}: You have received concise, pre-filtered intelligence reports for [Asset Name] ([Asset Ticker]) as of [Analysis Date]. Your task is to look for confluence or divergence between these reports to construct distinct lines of reasoning for either a bullish or bearish stance.

\textbf{Input Data}:

You will be provided with a collection of structured summaries from different analytical perspectives.

1.  \textbf{Fundamental Analysis Summary}:[fundamental\_output]

2.  \textbf{Technical Analysis Summary}:[technical\_output]

3.   \textbf{Narrative Analysis Summary}:[news\_output]

\textbf{Task}:

1.  \textbf{Synthesize}: Review all provided analysis summaries. Identify distinct lines of reasoning by connecting facts across different reports. A single fact is not an argument; a synthesis of facts is.

2.  \textbf{Formulate Arguments}: For each distinct line of reasoning, construct a structured investment argument. An argument can be supported by evidence from one or more of the analytical summaries.

3.  \textbf{Assign Stance}: Clearly define each argument as either bullish or bearish.

Constraints:

1. The output must be a list of structured JSON objects.

2. Each argument must be a self-contained, logical thesis.

3. The `rationale` should be your own synthesized conclusion, not a direct copy of the input.

4. The `evidence` must be a direct quote or a close, factual paraphrase of the key observations from the input summaries that support your rationale.

\textbf{Output Format}:

Produce a valid JSON array of argument objects. Each object must conform to the following schema:

\{{

    "p": 1 for bullish. -1 for bearish,
    
    "a": "argument analytical rationale",
    
    "e": "argument evidence"

\}}
}

\end{tcolorbox}

\subsection{Dataset composition details}
\label{appendix::dataset}
\subsubsection{Stock pool details}
\begin{itemize}
    \item \textbf{SSE 50}: This index represents the 50 largest and most liquid A-share stocks on the Shanghai Stock Exchange. It captures the performance of leading blue-chip companies.
    
    \item \textbf{CSI 300}: This index is a broad market benchmark for China's A-share market. It consists of the 300 largest and most liquid stocks from both the Shanghai and Shenzhen exchanges, offering a comprehensive representation of the large-cap segment.
    
    \item \textbf{CSI 500}: This index tracks the performance of the next 500 largest and most liquid A-share stocks after the constituents of the CSI 300. It primarily reflects the behavior of mid and small-cap companies, providing a distinct universe from the large-cap focused indices.
\end{itemize}

\subsubsection{Dataset construction details}
\paragraph{Fundamental Modality}
\begin{itemize}
    \item Financial Statements, including Income Statement, Balance Statement, CashFlow Statement
    \item Valuation Metrics, including Price-to-Earnings Ratio, Price-to-Book Ratio, Price-to-Cash Flow Ratio, Price-to-Sales Ratio, and dividend Ratio. 
\end{itemize}
\paragraph{News Modality}
\begin{itemize}
    \item News collected from Wind \footnote{\url{https://www.wind.com.cn/}}.
    \item Financial Report Collected from EastMoney \footnote{\url{https://data.eastmoney.com/report/}}.
\end{itemize}
\paragraph{Technical Modality}
\begin{itemize}
\item \textbf{Price and Volume Series}: Daily, weekly, and monthly adjusted close prices,  trading volumes, and market cap.
\item \textbf{Technical Indicators}:
\begin{itemize}
\item Trend-following: Daily, weekly, and monthly moving averages.
\item Momentum: Daily, weekly, and monthly Moving Average Convergence Divergence (MACD) and Relative Strength Index (RSI).
\item Volatility: Daily, weekly, and monthly Bollinger Bands (BOLL).
\item Oscillators: Daily, weekly, and monthly Stochastic Oscillator (KDJ).
\end{itemize}
\item \textbf{Contextual Metadata}: The primary industry classification for each stock.
\item \textbf{Benchmark Data}: Daily price series for the corresponding industry indices.
\end{itemize}

\subsection{Performance Metrics Details}
\label{appendix::metrics}

We assess the performance of a predictive model denoted by $f$. At each time step $t$, for a universe of $N_t$ assets, the model generates a vector of predictive scores $\mathbf{s}_t \in \mathbb{R}^{N_t}$. The efficacy of these scores is measured against the vector of subsequent asset returns $\mathbf{r}_t \in \mathbb{R}^{N_t}$. Unless specified otherwise, all statistical moments such as expectation $\mathbb{E}[\cdot]$, variance $\operatorname{Var}[\cdot]$, and covariance $\operatorname{Cov}[\cdot,\cdot]$ are computed cross-sectionally over the asset universe at time $t$.

The fundamental measure of predictive accuracy is the daily cross-sectional Pearson correlation between the scores and returns.
\begin{equation}
    \rho_t = \frac{\operatorname{Cov}(\mathbf{s}_t, \mathbf{r}_t)}{\sqrt{\operatorname{Var}(\mathbf{s}_t) \operatorname{Var}(\mathbf{r}_t)}}
\end{equation}

For portfolio-based metrics, we simulate a strategy where the daily portfolio return, $R^p_t$, is constructed from the factor scores $\mathbf{s}_t$. The cumulative wealth over a period of $T$ time steps is given by $W_T = \prod_{t=1}^{T} (1 + R^p_t)$. Let $A$ be the number of trading periods in a year, e.g., $A=252$ for daily data, and let $r_{t,f}$ be the risk-free rate.

\paragraph{Information Coefficient}
The Information Coefficient (IC) measures the time-series average of the daily cross-sectional correlations, quantifying the mean predictive power of the factor.
\begin{equation}
    \text{IC} = \mathbb{E}_t[\rho_t]
\end{equation}

\paragraph{Information Ratio}
The Information Ratio (IR) evaluates the consistency of the factor's predictive power. It is the time-series mean of the daily correlations scaled by their standard deviation, thus rewarding stable performance.
\begin{equation}
    \text{IR} = \frac{\mathbb{E}_t[\rho_t]}{\operatorname{Std}_t(\rho_t)}
\end{equation}

\paragraph{Rank Information Coefficient}
The Rank Information Coefficient (Rank IC) is the time-series average of the daily Spearman rank correlations. It assesses the model's ability to correctly order assets by predicted performance and is robust to outliers and non-linear monotonic relationships.
\begin{equation}
    \text{Rank IC} = \mathbb{E}_t[\operatorname{Corr}(\operatorname{rank}(\mathbf{s}_t), \operatorname{rank}(\mathbf{r}_t))]
\end{equation}

\paragraph{Rank Information Ratio}
The Rank Information Ratio (Rank IR) measures the stability of the factor's ordinal predictive power over time.
\begin{equation}
    \text{Rank IR} = \frac{\mathbb{E}_t[\operatorname{Corr}(\operatorname{rank}(\mathbf{s}_t), \operatorname{rank}(\mathbf{r}_t))]}{\operatorname{Std}_t(\operatorname{Corr}(\operatorname{rank}(\mathbf{s}_t), \operatorname{rank}(\mathbf{r}_t)))}
\end{equation}

\paragraph{Annualized Return}
The Annualized Return (AR) represents the total return of the factor-induced portfolio scaled to a one-year period.
\begin{equation}
    \text{AR} = A \cdot \mathbb{E}_t[R^p_t]
\end{equation}

\paragraph{Maximum Drawdown}
The Maximum Drawdown (MDD) quantifies the largest peak-to-trough decline in the portfolio's cumulative wealth, serving as a critical measure of tail risk.
\begin{equation}
    \text{MDD} = \max_{t \le T} \left( 1 - \frac{W_t}{\max_{u \le t} W_u} \right)
\end{equation}

\paragraph{Sharpe Ratio}
The annualized Sharpe Ratio (SR) measures the risk-adjusted excess return of the portfolio. It is defined as the annualized ratio of the average excess return to the standard deviation of excess returns.
\begin{equation}
    \text{SR} = \frac{\mathbb{E}_t[R^p_t - r_{t,f}]}{\operatorname{Std}_t(R^p_t - r_{t,f})} \cdot \sqrt{A}
\end{equation}

\subsection{Baseline Details}
\label{appendix::baseline}
We conduct experiments on various SOTA baselines:

\paragraph{Machine Learning Approaches} 
\begin{itemize}
    \item LightGBM~\cite{lgbm}:  A high-efficiency, leaf-wise gradient boosting decision tree framework by Microsoft research. We utilize the offical python package~\footnote{\url{https://github.com/microsoft/LightGBM}} to implement it.
\end{itemize}

\paragraph{Deep Learning Approaches}
\begin{itemize}
    \item \textbf{DTML}: DTML \citep{yoo2021accurate} presents an innovative attention-based architecture designed to capture intricate inter-stock correlations for investment decision-making. Our implementation is based on the official publicly available code\footnote{\url{https://github.com/ceteris11/DTML}}.
    \item \textbf{FactorVAE}: We include FactorVAE \citep{duan2022factorvae}, an influential probabilistic dynamic factor model that leverages a variational autoencoder framework. The implementation is adapted from the open-source repository\footnote{\url{https://github.com/x7jeon8gi/FactorVAE}}.
    \item \textbf{MASTER}: MASTER \citep{li2024master} is a powerful Transformer-based model for stock price forecasting. It adeptly models both contemporaneous and temporal stock correlations while integrating market-wide information to guide feature selection. We utilize the authors' provided implementation\footnote{\url{https://github.com/SJTU-DMTai/MASTER}}.
\end{itemize}

\paragraph{Agent-Based Approaches}
\begin{itemize}
    \item \textbf{SEP}: SEP \citep{koa2024learning} introduces a novel agent-based approach where a verbal self-reflective agent, trained with Proximal Policy Optimization, learns to generate explainable predictions for individual stocks. We build our tests upon the official open-source code\footnote{\url{https://github.com/koa-fin/sep}}.
    \item \textbf{TradingAgents}: TradingAgents \citep{xiao2025tradingagents} offers a sophisticated multi-agent framework that constructs investment portfolios by simulating the collaborative dynamics within trading firms. Our experiments leverage the publicly accessible implementation\footnote{\url{https://github.com/TauricResearch/TradingAgents}}.
    \item \textbf{R\&D-Agent(Q)}:R\&D-Agent(Q) \citep{li2025rdagentquant} is a pioneering LLM-driven framework designed for collaborative factor model development in quantitative finance. For this baseline, we employ the official implementation provided by the authors\footnote{\url{https://github.com/microsoft/RD-Agent}}.
\end{itemize}

\begin{figure}
    \centering
    \includegraphics[width=0.55\linewidth]{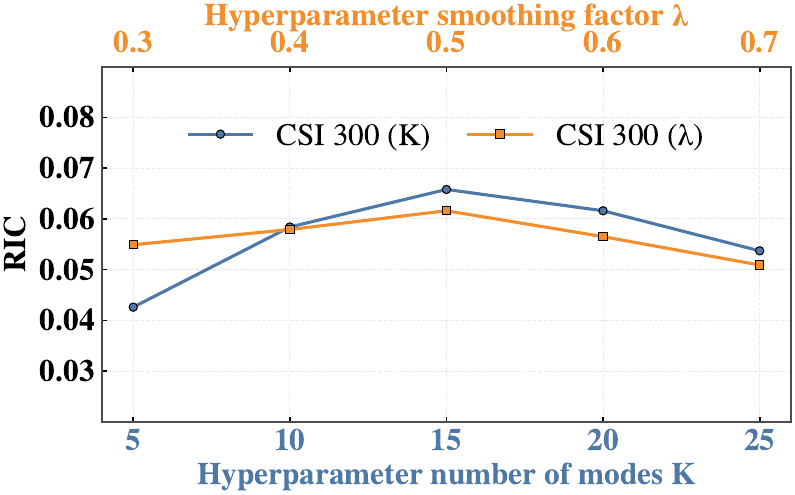}
    \caption{\MYMETHOD~exhibits a moderate sensitivity to changes in hyperparameters on RIC.}
    \label{fig:ParameterSensitivityExperiments_ric}
\end{figure}

\begin{figure}
    \centering
    \includegraphics[width=0.55\linewidth]{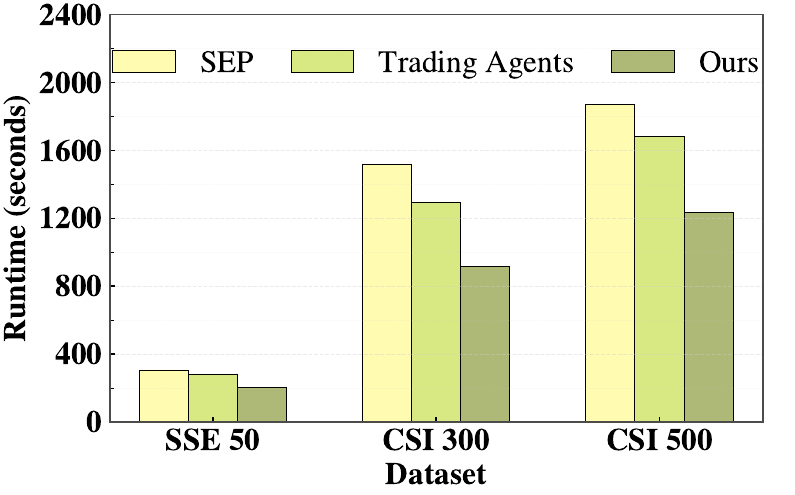}
    \caption{Average Daily running time on \MYMETHOD~and two LLM-based SOTA baselines.}
    \label{fig:runtime}
\end{figure}

\subsection{Detailed examples of effective modes}
\label{appendix::modes_details}
\subsubsection{Classification of modes}
\begin{itemize}
    \item Long-term effective:Across both bull and bear markets, the selected mode exhibits positive values in Equation~\ref{eq:perf_update} for over 65\% of the observed periods.
    \item Bull-market effective: the selected mode is not long-term effective, and exhibits positive values in Equation~\ref{eq:perf_update} for over 70\% of the bull-market periods.
    \item Bear-market effective: the selected mode is not long-term effective and bull-market effective, and exhibits positive values in Equation~\ref{eq:perf_update} for over 70\% of the bear-market periods.
    \item Non-effective: None of the above.
\end{itemize}

\subsubsection{Long-term effective modes}
\begin{tcolorbox}[
  enhanced,                   
  breakable,                  
  colback=EFE9E3,             
  colframe=C9B59C,     
  boxrule=1.0pt,              
  arc=2mm,                    
  left=4pt, right=4pt,        
  top=4pt, bottom=4pt,
  width=\linewidth,           
  title={\bfseries Modes example \& summary},
  fonttitle=\bfseries
]
{\ttfamily\small\justifying
\textbf{Modes examples}

\{{"converted\_arguments": [
  
   \{{"code": "000001.SZ",
      
      "name": "Ping An Bank",
      
      "date": "2024-01-30",
      
      "argument": \{{
      
        "p": 1,
        "a": "The stock is significantly undervalued with substantial potential for mean reversion.",
        "e": "Based on a relative valuation and historical quantile analysis framework, the current P/E ratio of 3.80 is only above the 2.8th percentile of its historical data, far below the median of 8.84. Similarly, the P/B ratio of 0.40 is at the 2.8th percentile, below the median of 0.82, indicating significant room for valuation recovery."
        
      \}}\}},
      
    \quad
    
    \{{"code": "000895.SZ",
      
      "name": "Henan Shuanghui Investment \& Development",
      "date": "2024-06-25",
      
      "argument": \{{
      
        "p": 1,
        
        "a": "The inventory management strategy is poised to enhance profitability.",
        
        "e": "Based on the company's statement on optimizing frozen goods inventory to maximize returns, combined with industry data showing a month-over-month contraction in hog production, this cost control optimization is expected to directly improve the gross profit margin."
      \}}\}},
          
    \quad

    \{{
      "code": "603259.SH",
      
      "name": "WuXi AppTec",
      
      "date": "2024-07-08",
      
      "argument": \{{
        "p": 1,
        
        "a": "Utilizing a model that links buyback activities with shareholder confidence, the cancellation of stock options sends a positive signal.",
        
        "e": "WuXi AppTec's announcement on July 1st to cancel 31,500 stock options, viewed in the context of over 1,600 A-share companies conducting nearly 100 billion RMB in buybacks in the first half of the year, reflects the management's proactive stance on supporting the company's valuation."
        
      \}}\}},
          
    \quad
    
    \{{
      "code": "000725.SZ",
      
      "name": "BOE Technology Group A",
      
      "date": "2025-05-27",
      
      "argument": \{{
      
        "p": -1,
        
        "a": "Intensifying industry competition and capital expenditure pressures are likely to weigh on short-term performance.",
        
        "e": "A competitive landscape analysis reveals that competitors like HKC and Tianma have signed new display projects totaling 72 billion RMB for mass production. This acceleration in industry capacity expansion means BOE Technology will face heightened challenges from price competition and market integration."
        
      \}}\}},
          
    \quad
    
    \{{"code": "601138.SH",
      
      "name": "Foxconn Industrial Internet",
      
      "date": "2025-04-12",
      
      "argument": {
      
        "p": 1,
        
        "a": "Earnings growth is the primary driver for stock price appreciation.",
        
        "e": "According to a performance-driven analytical framework, the company's net profit for the first quarter grew by 24\%-27\% year-over-year, with revenue increasing by 34\%-35.6\%, setting a new record for the period. This high-growth data substantiates expectations for an upward trajectory in the stock price."
        
      \}}\}}
    
  ],\}}

\textbf{Modes summaries}

The overarching mode of thought is a multi-faceted, evidence-based reasoning process. It synthesizes diverse data from quantitative metrics and corporate actions to industry trends—through specific analytical frameworks to construct a structured investment thesis on a stock's future performance.
}
\end{tcolorbox}
\subsubsection{Bull-market effective modes}
\begin{tcolorbox}[
  enhanced,                   
  breakable,                  
  colback=EFE9E3,             
  colframe=C9B59C,     
  boxrule=1.0pt,             
  arc=2mm,                    
  left=4pt, right=4pt,        
  top=4pt, bottom=4pt,
  width=\linewidth,           
  title={\bfseries Modes example \& summary},
  fonttitle=\bfseries
]
{\ttfamily\small\justifying
\textbf{Modes examples}

\{{"converted\_arguments": [
  
   \{{"code": "000063.SZ",
      
      "name": "ZTE Corporation",
      
      "date": "2024-01-30",
      
      "argument": \{{
      
        "p": -1,
        "a": "The trend has deteriorated on a weekly basis, confirming the continuation of a mid-term downtrend.",
        
        "e": "The weekly closing price of 20.78 has fallen below the Bollinger Band middle rail of 25.73, the RSI has dropped sharply to 30.9, and the negative MACD value has widened to -0.62."
        
      \}}\}},
      
    \quad
    
    \{{"code": "688256.SH",
      
      "name": "Cambricon Technologies",
      "date": "2025-07-30",
      
      "argument": \{{
      
        "p": 1,
        
        "a": "A monthly-level breakout above the upper Bollinger Band, coupled with an overbought RSI, indicates strong upward momentum.",
        
        "e": "Based on trend continuation logic, the recent monthly closing price of 1492.49 is significantly above the upper Bollinger Band of 1127.98, and the monthly RSI has reached 84.26, confirming a clear breakout pattern."
      \}}\}},
          
    \quad

    \{{
      "code": "600519.SH",
      
      "name": "Kweichow Moutai",
      
      "date": "2024-05-06",
      
      "argument": \{{
        "p": 1,
        
        "a": "The stock is exhibiting a strong short-term technical breakout, suggesting the price is likely to continue rising.",
        
        "e": "Based on a bullish moving average alignment framework, the daily closing price has stayed above all short-term moving averages (5-day, 10-day, 20-day) for three consecutive days, and the 5-day MA has crossed above the 10-day and 20-day MAs, indicating strong short-term upward momentum."
        
      \}}\}},
          
    \quad
    
    \{{
      "code": "688981.SH",
      
      "name": "SMIC",
      
      "date": "2025-07-30",
      
      "argument": \{{
      
        "p": 1,
        
        "a": "The short-term stock price is poised to benefit from institutional capital inflows and positive market sentiment.",
        
        "e": "Margin financing net purchases on July 29th reached a two-month high of 164 million RMB, and the margin financing balance ratio rose to 4.01\% on July 30th. Concurrently, recommendations from Minsheng Securities and Everbright Securities as a top pick for August signal strong institutional optimism.."
        
      \}}\}},
          
    \quad
    
    \{{"code": "603986.SH",
      
      "name": "GigaDevice Semiconductor",
      
      "date": "2025-04-12",
      
      "argument": {
      
        "p": 1,
        
        "a": "The stock price is likely to increase, as inferred from financing activities and capital flow dimensions.",
        
        "e": "A net margin financing purchase of 65.38 million RMB on June 30th marked the fourth consecutive day of net purchases, indicating active capital inflow from the market and reflecting strengthening investor confidence."
        
      \}}\}}
    
  ],\}}

\textbf{Modes summaries}

This thinking mode prioritizes technical analysis and market sentiment. It interprets chart patterns, momentum indicators (RSI, MACD), and capital flows (margin financing) as direct predictors of short-term price movements, inferring investor conviction and forecasting trend continuation or reversal based on these signals.

}
\end{tcolorbox}
\subsubsection{Bear-market effective modes}
\begin{tcolorbox}[
  enhanced,                   
  breakable,                  
  colback=EFE9E3,             
  colframe=C9B59C,     
  boxrule=1.0pt,              
  arc=2mm,                    
  left=4pt, right=4pt,        
  top=4pt, bottom=4pt,
  width=\linewidth,           
  title={\bfseries Modes example \& summary},
  fonttitle=\bfseries
]
{\ttfamily\small\justifying
\textbf{Modes examples}

\{{"converted\_arguments": [
  
   \{{"code": "600941.SH",
      
      "name": "China Mobile",
      
      "date": "2024-01-30",
      
      "argument": \{{
      
        "p": 1,
        "a": "An attractive dividend yield provides downside protection and enhances investment appeal.",
        
        "e": "Utilizing a yield analysis framework, the dividend yield stands at 4.15\%, surpassing the industry average of 2.44\%. The company's abundant cash flow, with cash and equivalents consistently exceeding 3.3 trillion RMB and no interest-bearing debt, supports the sustainability of its dividend policy."
        
      \}}\}},
      
    \quad
    
    \{{"code": "601816.SH",
      
      "name": "Beijing-Shanghai High-Speed Railway",
      "date": "2024-06-27",
      
      "argument": \{{
      
        "p": 1,
        
        "a": "The implementation of a cash dividend may attract stable, yield-seeking capital inflows.",
        
        "e": "Based on a dividend event-driven analysis, the company's execution of a cash dividend of 1.12 RMB per 10 shares, with the record date completed on June 27th, highlights its prominent high-dividend characteristic."
      \}}\}},
          
    \quad

    \{{
      "code": "688981.SH",
      
      "name": "SMIC",
      
      "date": "2024-05-12",
      
      "argument": \{{
        "p": -1,
        
        "a": "Deteriorating cash flow combined with a lack of dividends diminishes investment attractiveness and may lead to valuation pressure.",
        
        "e": "Through P/CF and dividend yield analysis, the company's cash flow return is insufficient. The P/CF ratio of 14.55 is above its historical 36th percentile while cash and equivalents have declined. The 0\% dividend yield presents a clear disadvantage compared to the industry's average yield.."
        
      \}}\}},
          
    \quad
    
    \{{
      "code": "601012.SH",
      
      "name": "LONGi Green Energy",
      
      "date": "2025-04-24",
      
      "argument": \{{
      
        "p": -1,
        
        "a": "The balance sheet reveals growing financial pressure, which constrains potential stock price appreciation.",
        
        "e": "Cash and equivalents have decreased while total debt remains high, causing the debt-to-asset ratio to rise to 0.592. Persistently high levels of accounts receivable and inventory signal pressure on operational efficiency."
        
      \}}\}},
          
    \quad
    
    \{{"code": "600900.SH",
      
      "name": "China Yangtze Power",
      
      "date": "2025-08-27",
      
      "argument": {
      
        "p": 1,
        
        "a": "The stock price is likely to increase, as inferred from financing activities and capital flow dimensions.",
        
        "e": "Applying dividend and cash flow analysis, the current dividend yield of 3.41\% is above the industry's 2.67\%, and a P/CF ratio of 11.38 below its historical median indicates a reasonable valuation. An improving financial structure, seen in the reduced debt-to-asset ratio, enhances dividend sustainability."
        
      \}}\}}
    
  ],\}}

\textbf{Modes summaries}

This thinking mode prioritizes financial health and shareholder returns. It assesses a company's intrinsic value and risk profile by analyzing balance sheet strength, cash flow generation, and dividend policies, viewing a strong dividend and stable financials as key indicators of investment quality and downside protection.

}
\end{tcolorbox}

\end{document}